\documentclass[10pt,twocolumn,letterpaper]{article}
\pdfoutput=1
\usepackage{iccv}
\usepackage{times}
\usepackage{epsfig}
\usepackage{graphicx}
\usepackage{amsmath}
\usepackage{amssymb}
\usepackage{multirow}
\usepackage{color}
\usepackage{lipsum}

\usepackage{algorithm,caption}

\usepackage{mathtools}

\usepackage[hyperindex,colorlinks,bookmarks=false,pdftex]{hyperref}

 \iccvfinalcopy 

\def\iccvPaperID{9} 

\ificcvfinal\fi
\begin{document}
\title{On Generalizing Detection Models for Unconstrained Environments}

\author{Prajjwal Bhargava\\
{\tt\small prajjwalin@protonmail.com}}

\maketitle

\begin{abstract}
   Object detection has seen tremendous progress in recent years. However, current algorithms don't generalize well when tested on diverse data distributions. We address the problem of incremental learning in object detection on the India Driving Dataset (IDD). Our approach involves using multiple domain-specific classifiers and effective transfer learning techniques focussed on avoiding catastrophic forgetting. We evaluate our approach on the IDD and BDD100K dataset. Results show the effectiveness of our domain adaptive approach in the case of domain shifts in environments. 
\end{abstract}

\section{Introduction}

Object detection has been a widely studied task in computer vision. It is focussed upon classifying objects present in an image and then regressing bounding boxes over the localized proposals. We have seen remarkable results with CNN based models\cite{NIPS2012_4824} on the COCO dataset\cite{DBLP:journals/corr/LinMBHPRDZ14} \cite{DBLP:journals/corr/abs-1805-09300} \cite{DBLP:journals/corr/abs-1904-08189} \cite{DBLP:journals/corr/abs-1901-01892} \cite{DBLP:journals/corr/abs-1904-11492}. Recently, \cite{DBLP:journals/corr/abs-1906-02659} showed that when commonly used detectors are evaluated on nonstandard settings of objects in an environment, they tend to provide unusual predictions. This is also applicable for autonomous navigation systems operating in unstructured environments (e.g drivable areas except roads etc.) as well. Current detection methods don't generalize well when they encounter diverse environmental conditions. 

We witness variety of environmental conditions when it comes to driving such as weather changes, dynamic changes in the surrounding environment, etc. Current detectors have been tested on data obtained from structured environments which are often not representative of real-world conditions. As a result of which, the need for data obtained from nonstandard sources is felt the most for data-driven algorithms to improve and test their generalizing capabilities.

\begin{figure}
\centering
      \resizebox{0.47\textwidth}{!}{%
    \setlength{\fboxsep}{0pt}
      \fbox{\includegraphics[width=0.16\textwidth,height=0.11\textwidth]{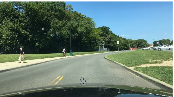}}
      \fbox{\includegraphics[width=0.16\textwidth,height=0.11\textwidth]{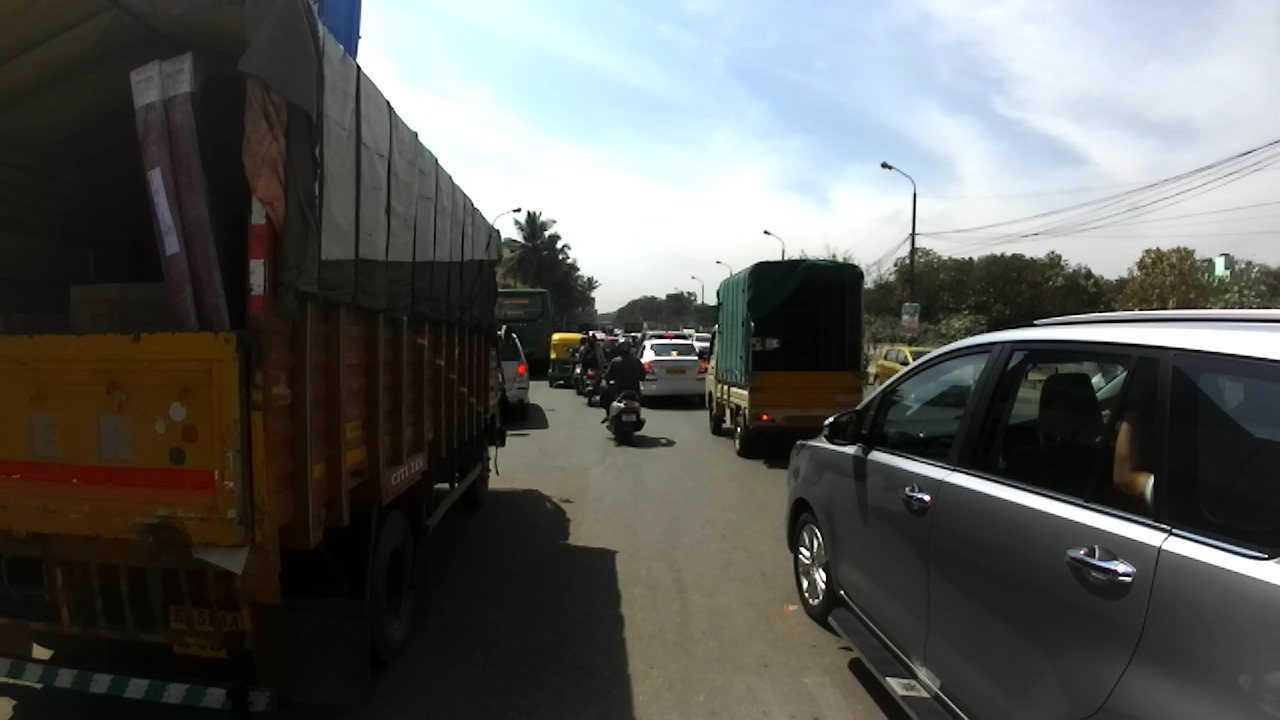}}
      \fbox{\includegraphics[width=0.16\textwidth,height=0.11\textwidth]{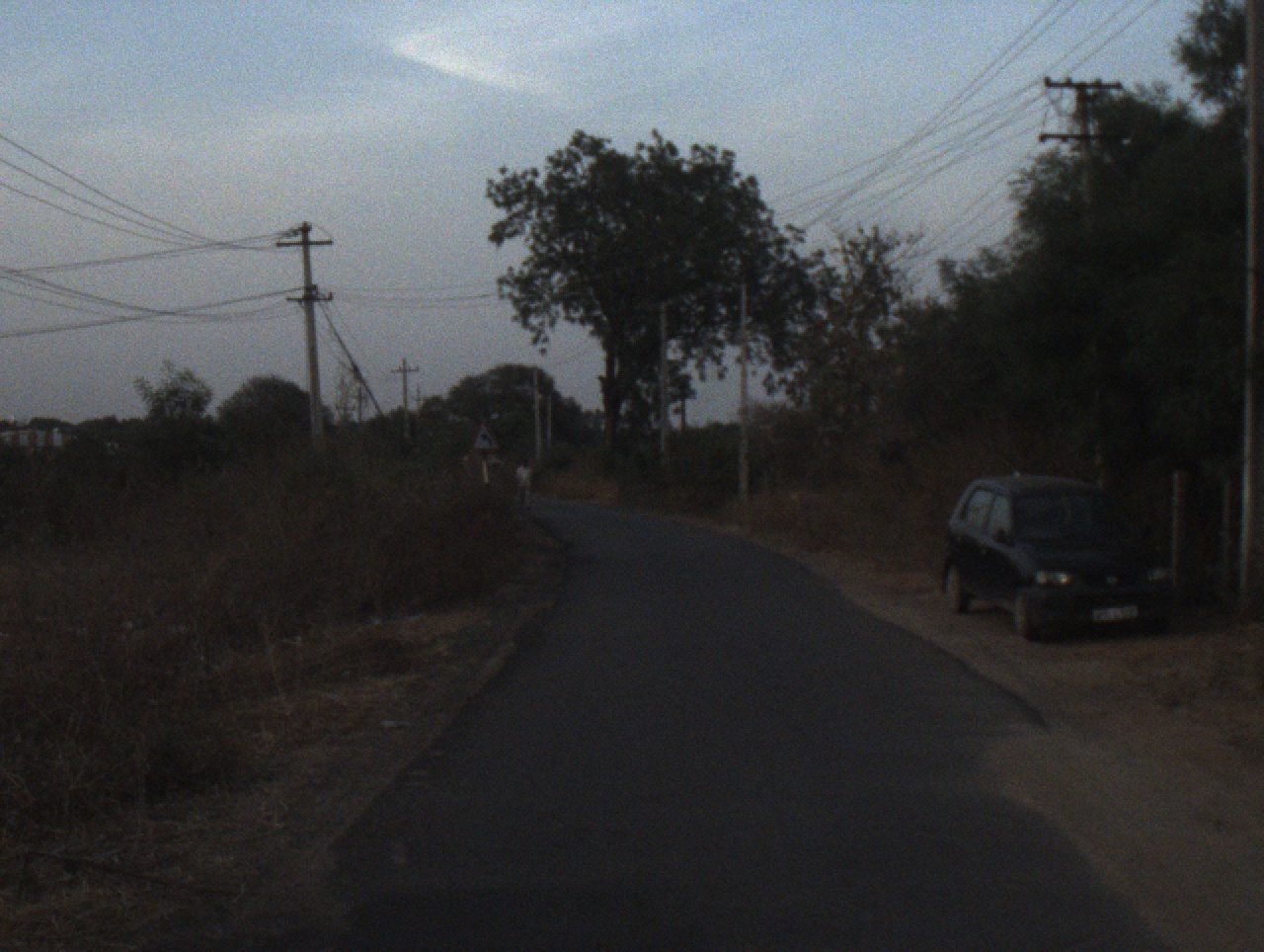}}
      }\\
\caption{\textbf{Illustration of datasets for autonomous driving used in this work:} Leftmost image is taken from:  \textit{BDD100K}\cite{Cordts2015Cvprw} and the other two are from \textit{IDD}\cite{DBLP:journals/corr/abs-1811-10200}. Even though datasets for autonomous navigation aim to include diverse features such as illumination, various styles etc. IDD is very different in regards to vehicle density,road boundaries,diverse ambient conditions.}
\label{fig:ds_example}
\vspace{-10pt}
\end{figure}

Autonomous navigation algorithms must perform well on multiple domains especially the ones with corner cases for safety purposes. Most importantly, we want to be able to learn from a large standard data distribution to efficiently learn features in an embedding space and learn progressively from domain-specific data without having access to earlier used data. 

In this paper, we address the problem of incremental learning and domain adaptation to some extent for object detectors to improve generalizing capabilities. Specifically, we tackle the problem of adapting from a standard data distribution to data obtained from the unstructured environment. We also provide baseline results on IDD and BDD100K for object detection task to compare our proposed methods.\footnote{Code for this work can be found \href{https://github.com/prajjwal1/autonomous-object-detection}{here}} 

\begin{figure*}
\centering
\includegraphics[width=1.0\linewidth]{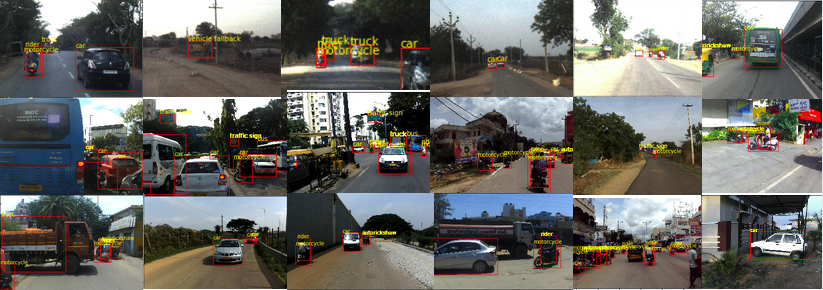}
\caption{\textbf{Sample of predictions from our baseline model} trained on non HQ image set from IDD. As evident, images in IDD are highly diverse. In some cases, environments are highly unstructured, while in some cases objects of interest are occluded or far off.}
\label{fig:Predictions_from_baseline_model}
\vspace{-2mm}
\end{figure*}

\section{Related work}

\textbf{Object Detection:}
Region proposal based methods introduced in \cite{DBLP:journals/corr/GirshickDDM13} have been widely used as object detectors. It made use of selective search to reduce the number of bounding boxes. Spatial Pyramid Pooling Nets \cite{DBLP:journals/corr/HeZR014} could generate a fixed-length representation in a dynamic manner irrespective of image scale. Fast RCNNs \cite{DBLP:journals/corr/Girshick15} made use of regression for bounding box predictions. \cite{NIPS2015_5638} made use of RPNs and introduced anchor boxes to deal with different aspect ratios and scales. 
SSD \cite{DBLP:journals/corr/LiuAESR15} method runs a CNN on input image only once and calculates a feature map that doesn't require proposal generation steps. Stereo RCNNs \cite{DBLP:journals/corr/abs-1902-09738} extends the use of Faster RCNN with stereo images for 2D and 3D bounding box predictions. It is a region proposal based network that works without the need for point clouds. Our approach can also be extended for 3D object detection similarly but we still lack the diversified ground truth data (such as 3D bounding box coordinates or Lidar point clouds obtained from unconstrained environments) for 3D detections. 

\textbf{Learning from multiple distributions:}
The concept of making generalizable deep learning models has been widely studied. This often involves retaining what the model has learned in the past and performing incremental learning on multiple domains. \cite{inproceedings} used a GAN\cite{NIPS2014_5423} to
approximate the feature distribution in the source domain. \cite{DBLP:journals/corr/LiH16e} \cite{DBLP:journals/corr/abs-1808-06281} addressed the task of incremental learning with architectures that inhibit loss of learned knowledge. \cite{44873} made use of a larger network to train a smaller network to generate close predictions. \cite{DBLP:journals/corr/CourtyFTR15} treated the task of domain adaptation as an optimal transport problem. 

\section{Preliminaries}
\textbf{Faster RCNN:} It takes an RGB image as an input. The model consists of a feature extractor followed by a feature pyramid network (FPN) and region proposal network (RPN) for generating region proposals which are then used to detect objects. RPNs are more efficient than selective search. They perform a ranking of anchor boxes to reduce their number and propose those which most likely contain an object. Image features are generated by a backbone network which is then fed to an RPN along with images and targets for generating proposals. After RPN, we get proposed regions with different sizes. Region of Interest (ROI) classifier predicts the category label obtained by using ROI Pooling. RPN can output differently sized regions. ROI Pooling can simplify the problem by reducing the feature maps into the same size. The loss is the sum of classification and regression loss defined as:
\begin{equation}
L_{d e t} = L_{c l s} + L_{r e g}   
\end{equation}
We refer readers to \cite{NIPS2015_5638} for further details about model architecture.

\label{sec:dadet}
\begin{figure*}
\centering
\includegraphics[width=0.9\linewidth,trim=1 1 1 1,clip]{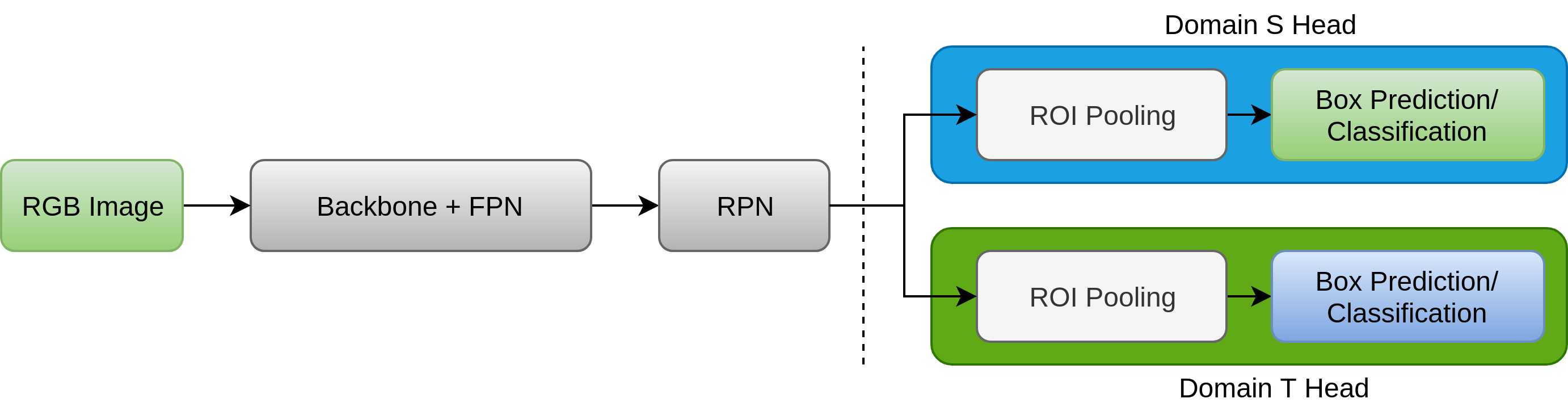}
\caption{\label{fig:model} \textbf{Overview of our architecture} We use multiple classifier heads to perform domain-specific predictions. Rest of the network shares the same weights to learn common representations}
\label{fig:fig_model}
\vspace{-2mm}
\end{figure*}

\section{Baseline Model}
We use a region proposal based approach for the baseline model. For feature extraction which is used by RPN, we use ResNet50\cite{DBLP:journals/corr/HeZRS15} followed by FPN pretrained on COCO. Linear layers after ROI Head were adjusted as per the number of classes.
 Our RPN generates 5 x 3 anchors per spatial location with 5 different sizes and 3 different aspect ratios. We used random horizontal flipping for augmenting the input data. Our baseline model is trained on a non HQ image set from IDD with batch size set to 4 for 5 epochs per camera orientation. It was optimized using SGD \cite{Sutskever:2013:IIM:3042817.3043064} with momentum and weight decay set to 0.9 and 0.00004 respectively. The learning rate was initially set to 0.001 with the Cyclical learning rate scheduler\cite{DBLP:journals/corr/Smith15a}. We use the same process for performing training on BDD100K. Results are shown in \autoref{tab::table_1} and \autoref{tab::table_2}.

 \begin{table}
\begin{center}
\begin{tabular}{|l|c|c|}
\hline
IoU & Area & mAP(\%)\\
\hline\hline\hline
 0.50  & all & 31.57 \\
 0.75 & all & 18.9 \\
 0.50:0.95 & medium & 16.44 \\
 0.50:0.95 & large & 36.83 \\ 
 0.50:0.95 & all & 18.45 \\ [1ex]
\hline
\end{tabular}
\end{center}
\caption{Quantitative results from baseline model reported on validation set of \textbf{IDD}}
\label{tab::table_1}
\end{table}

 \begin{table}
\begin{center}
\begin{tabular}{|l|c|c|}
\hline
IoU & Area & mAP(\%)\\
\hline\hline\hline
 0.50  & all & 45.7 \\
 0.75 & all & 18.2 \\
  0.50:0.95 & medium & 32.0 \\
 0.50:0.95 & large &  40.63 \\ 
 0.50:0.95 & all & 22.65  \\ [1ex] 
\hline
\end{tabular}
\end{center}
\caption{Quantitative results from baseline model reported on validation set of \textbf{BDD100K}}
\label{tab::table_2}
\end{table}

\section{Incremental Learning}
The following section contains a description of training methodology and proposed transfer learning techniques aimed at minimizing catastrophic learning while adapting to target $T$ data distribution. 

Our task is to perform incremental learning on multiple diverse data distributions. The network initially learns the weights from a standard data distribution and the proposed techniques help in performing domain adaptation while remaining consistent with the already learned information. Once trained on one distribution, we don't require already used data. 

\subsection{Domain specific heads} We make use of two ROI heads which are combined with the common backbone and RPN for generating domain-specific predictions.  We can also have more than two ROI heads depending upon the number of target domains we want to adapt to. The weights of RPN and feature extractor are shared across all domain-specific classifiers. Weight sharing allows the network to learn common features with the proposed techniques across all domains without any increment in the number of parameters. Domain-specific heads also help in cases where classes don't overlap in both distributions, as in this case.

After the addition of ROI Head to the baseline model, we train the head on $T$ to learn domain-specific weights. This is followed by progressive training of other components of the network to avoid catastrophic forgetting as proposed in \cite{howard2018universal} to learn domain invariant features.

\subsection{Discriminative finetuning}
We use different learning rates to train different layers of our network. As shown in \cite{NIPS2014_5347}, different layers of the network are responsible for capturing different types of information. 
Discriminative finetuning allows us to set the rate at which these different components of the network learn. Since the weights of the backbone and RPN are being shared for all tasks, we want to inhibit the loss of learned information. We use a higher learning rate for domain-specific components and a lower learning rate for components whose weights are being shared. Specifically, we require a lower learning rate for the backbone and RPN since feature extraction and generation of region proposals are common tasks across all domains and a higher learning rate for domain-specific ROI Heads. A general SGD update of a model's parameters $\theta$ at time step $t$ looks like:
\begin{equation}
\theta_{t}=\theta_{t-1}-\eta \cdot \nabla_{\theta} J(\theta)    
\end{equation}
where $\eta$ denotes learning rate and $\nabla_{\theta} J(\theta)$ denotes gradient with respect to model's objective function. We split model's parameters $\theta$ into $\left\{\theta^{1}, \ldots, \theta^{L}\right\}$ where $\theta^{l}$ contains parameters of the model at the $l$-th layer and $L$ denotes the total number of layers of our network. The SGD update then becomes:
\begin{equation}
\theta_{t}^{l}=\theta_{t-1}^{l}-\eta^{l} \cdot \nabla_{\theta^{l}} J(\theta)
\end{equation}

\subsection{Gradual unfreezing}
Training the entire model on a different domain at once leads to catastrophic forgetting, which means the model adapts itself to the target domain on which it is being tuned compromising the performance on source domain on which it was trained. We overcome this issue by gradually unfreezing the components of the network with discriminative finetuning. We freeze all the components initially and unfreeze the domain ROI $T$ Head which is fine-tuned until convergence followed by progressive unfreezing and finetuning of FPN and RPN.

\subsection{Cyclical Learning Rate}
We optimize our network using the Cyclical learning rate (CLR) as proposed in \cite{smith2017cyclical}. Instead of having a gradually decreasing learning rate, as the training converges, we use CLR which cycles the learning rate between lower and upper bound. CLR helps in oscillating towards a higher learning rate wherever necessary. It prevents the network from converging at some poor local minima in loss landscape. We make use of triangular variation for our experiments.

\section{Experiments}
In this section, we evaluate our proposed approach on two diverse datasets. One dataset denotes structured environments and the other one denotes unstructured and unconstrained environments to which we want to adapt. The later one simulates high traffic density, rural areas with no proper roads, classes usually not seen in other datasets posing a much harder task for current object detection models. 

\subsection{Datasets}

\textbf{IDD: } We use IDD for target adaptation tasks. It provides data for object detection in two resolutions. The non HQ set consists of 27072 images taken from 5 different orientations of the camera with two resolutions 964x1280 and 1080x1920. The HQ set consists of 14722 images with two resolutions 720x1280 and 1080x1920. There are 15 classes for this task. Note that we only perform training and evaluation on the non HQ set of IDD. Results can be further improved if high res images from the HQ set are used to train the components of the network. The validation set consists of 10,225 high-resolution images. 

\textbf{Berkeley Deep Drive:} 
We use BDD100K \cite{DBLP:journals/corr/abs-1805-04687} to denote data distribution obtained from structured environments. We only use the images and their respective ground truths for the detection task. There are 69863 images in train and 10000 in the validation set. We trained our proposed model over 12 classes.  

\subsection{Training methodology}
The proposed architecture has been shown in \autoref{fig:fig_model}. This architecture is based on Faster RCNN. The backbone is a ResNet50 pretrained on COCO. We use a batch size of 16. We use the same baseline model with an additional ROI Head. We obtain four feature maps from the batch of images obtained by intermediate layers of backbone to perform multi-scale ROI aligning. These feature maps are shared across all components. The obtained feature maps are then fed to an RPN for generating region proposals followed by domain-specific ROI pooling and prediction layers. While training and inference, only the designated ROI Head is used for the respective domain. This model is trained in an end to end strategy and inference can be performed in a regular manner with learned weights. 

\subsection{Results}
In the following section, we evaluate the effects of each of the mentioned techniques along with the effect of varying learning rates.
As per convention, we use BDD$\rightarrow$IDD to denote BDD100K as $S$ and IDD as $T$. Since we have more data collected from structured environments \cite{DBLP:journals/corr/abs-1803-06184} \cite{DBLP:journals/corr/abs-1805-04687} \cite{Geiger2012CVPR} , the results simulate learning from already existing data distributions to adapt to unstructured environment.

\textbf{Adding domain specific head}
Here we use the same baseline model for BDD. We add a domain-specific head as proposed in \autoref{fig:model}. In this case, we only perform finetuning of this head on $T$. Apart from the domain-specific head, the rest of the components of the network are kept frozen. 

By introducing the domain-specific head and training it for 5 epochs, we see a considerable performance on $T$ without any performance decrement on $S$. BDD$\rightarrow$IDD indicates that we use the baseline model trained on BDD100K with the specified method. While reporting for IDD$\rightarrow$BDD, we only change the domain-specific head, the rest of the network stays the same. The same model can achieve an mAP of 24.3\% on IDD and 45.7\% on BDD. Results are shown in \autoref{tab::table_3}

\textbf{Discriminative finetuning and Gradual unfreezing}
We use the same network and weights as in the previous step. Here, we introduce both techniques. In \autoref{tab::table_3}, active components denote those components of the network whose weights are being updated during the training process. We experiment with different learning rates with progressive addition of active components with different bounds of learning rate during each step.

\begin{table*}
  \begin{center}
  \begin{tabular}{|c|c|c|c|c|}
  \hline
    $S$ and $T$ & Epoch & Active components (with LR) & LR Range & mAP (\%) at specified epochs \\
        \hline 
        BDD$\rightarrow$IDD & 5 & +ROI Head(1e-3)&1e-3, 6e-3&24.3 \\
        IDD$\rightarrow$BDD & Eval &  & - &45.7\\
        \hline
            BDD$\rightarrow$IDD &  5,9  &  +RPN (1e-4)  & 1e-4, 6e-4 & 24.7, 24.9 \\
            IDD$\rightarrow$BDD & Eval &  +ROI head (1e-3)   & -         & 45.3, 45.0 \\
        \hline
            BDD$\rightarrow$IDD &  1,5,6,7   &  +RPN (1e-4) & 1e-4, 6e-3 & 24.3, 24.9, 24.9, 25.0 \\
            IDD$\rightarrow$BDD & Eval &     +ROI head (1e-3) & -         & 45.7, 44.8, 44.7, 44.7 \\
        \hline
            BDD$\rightarrow$IDD &  1,5,10   &  +ROI head (1e-3)  & 1e-4, 6e-3 & 24.9, 25.4, 25.9 \\
            IDD$\rightarrow$BDD & Eval &  +RPN (4e-4) +FPN(2e-4) & -         & 45.2, 43.9, 43.3 \\
        \hline
  \end{tabular}
  \end{center}
  
  \caption{Change in mAP with varying learning rates for different active components. Results reported on validation sets of $T$ (IoU=0.5)}
  \label{tab::table_3}
\end{table*}

Learning rate plays a very crucial role in determining the performance increment on $T$ and retention of learned information. Domain-specific components require a higher learning rate as compared to shared components. The learning rate range plays a crucial role since it determines the rate at which weights change in all active components. In our experiments, we found this range 0.0001-0.006 for the learning rate to work well. As experimental results show, with a little decrement in performance on $S$, our model retains near similar performance on $S$ after being trained on $T$. In some cases, we saw an increment in performance on $T$ while maintaining the same performance on $S$. This shows that these transfer learning techniques complement each other and are effective in inhibiting information loss while adapting to diverse target distributions. 

\section{Discussion}
In this paper, we use an incremental learning approach and demonstrate the effectiveness of our method on data obtained from unconstrained environments. The main motivation behind this work is to demonstrate the effectiveness of our approach and encourage further research into building detection systems that generalize well on uncommon data distributions which are well representative of diverse real-world conditions. These proposed approaches can also be extended to other computer vision tasks as well.

\section{Acknowledgement}
The author would like to thank Intel AI for providing access to AI Devcloud as part of the Student ambassador program.

{\small
\bibliographystyle{ieee}
\bibliography{egbib}
}

\end{document}